\documentclass[11pt,a4paper]{article}
\usepackage[whole]{bxcjkjatype}
\usepackage[nohyperref]{acl2018}
\usepackage{times}
\usepackage{latexsym}
\usepackage{algpseudocode}
\usepackage{graphicx}
\usepackage{algorithm}
\usepackage{bm}
\usepackage{CJKutf8}
\usepackage{comment}
\usepackage{url}

\aclfinalcopy 
\def\aclpaperid{49} 

\algtext*{EndFor}

\title{Graph-based Filtering of Out-of-Vocabulary Words \\ for Encoder-Decoder Models}

\author{Satoru Katsumata, Yukio Matsumura\thanks{\, Both authors equally contributed to the paper.} , Hayahide Yamagishi\footnotemark[1] \and Mamoru Komachi\\
  Tokyo Metropolitan University \\
  {\tt \{katsumata-satoru, matsumura-yukio, yamagishi-hayahide\}@ed.tmu.ac.jp,} \\
  {\tt komachi@tmu.ac.jp} 
  }

\date{}

\begin{document}
\maketitle
\begin{abstract}
Encoder-decoder models typically only employ words that are frequently used in the training corpus to reduce the computational costs and exclude noise.
However, this vocabulary set may still include words that interfere with learning in encoder-decoder models.
This paper proposes a method for selecting more suitable words for learning encoders by utilizing not only frequency but also co-occurrence information, which we capture using the HITS algorithm.\footnote{\fontsize{7.5pt}{0cm}\selectfont \url{https://github.com/Katsumata420/HITS_Ranking}}
We apply our proposed method to two tasks: machine translation and grammatical error correction.
For Japanese-to-English translation, this method achieves a BLEU score that is 0.56 points more than that of a baseline.
Furthermore, it outperforms the baseline method for English grammatical error correction, with an $\mathrm{F_{0.5}}$-measure that is 1.48 points higher.
\end{abstract}

\section{Introduction}
Encoder-decoder models \cite{sutskever2014} are effective in tasks such as machine translation (\citeauthor{cho2014}, \citeyear{cho2014}; \citeauthor{bahdanau2014}, \citeyear{bahdanau2014}) and grammatical error correction \cite{yuan-briscoe2016}. 
Vocabulary in encoder-decoder models is generally selected from the training corpus in descending order of frequency, and low-frequency words are replaced with an unknown word token \verb|<unk>|. 
The so-called out-of-vocabulary (OOV) words are replaced with \verb|<unk>| to not increase the decoder's complexity and to reduce noise. 
However, naive frequency-based OOV replacement may lead to loss of information that is necessary for modeling context in the encoder. 

This study hypothesizes that vocabulary constructed using unigram frequency includes words that interfere with learning in encoder-decoder models. 
That is, we presume that vocabulary selection that considers co-occurrence information selects fewer noisy words for learning robust encoders in encoder-decoder models.
We apply the hyperlink-induced topic search (HITS) algorithm to extract the co-occurrence relations between words. 
Intuitively, the removal of words that rarely co-occur with others yields better encoder models than ones that include noisy low-frequency words.

This study examines two tasks, machine translation (MT) and grammatical error correction (GEC) to confirm the effect of decreasing noisy words, with a focus on the vocabulary of the encoder side, because the vocabulary on the decoder side is relatively limited.
In a Japanese-to-English MT experiment, our method achieves a BLEU score that is 0.56 points more than that of the frequency-based method.
Further, it outperforms the frequency-based method for English GEC, with an $\mathrm{F_{0.5}}$-measure that is 1.48 points higher.

The main contributions of this study are as follows:
\begin{enumerate}
\item The simple but effective preprocessing method we propose for vocabulary selection improves encoder-decoder model performance.
\item 
This study is the first to address noise reduction in the source text of encoder-decoder models.
\end{enumerate}

\section{Related Work}
There is currently a growing interest in applying neural models to MT (\citeauthor{sutskever2014}, \citeyear{sutskever2014}; \citeauthor{cho2014}, \citeyear{cho2014}; \citeauthor{bahdanau2014}, \citeyear{bahdanau2014}; \citeauthor{wu2016}, \citeyear{wu2016}) and GEC (\citeauthor{yuan-briscoe2016}, \citeyear{yuan-briscoe2016}; \citeauthor{xie2016}, \citeyear{xie2016}; \citeauthor{ji2017}, \citeyear{ji2017}); hence, this study focuses on improving the simple attentional encoder-decoder models that are applied to these tasks.

In the investigation of vocabulary restriction in neural models, \citet{sennrich2016} applied byte pair encoding to words and created a partial character string set that could express all the words in the training data.
They increased the number of words included in the vocabulary to enable the encoder-decoder model to robustly learn contextual information.
In contrast, we aim to improve neural models by using vocabulary that is appropriate for a training corpus---not to improve neural models by increasing their vocabulary.

\citet{jean2015} proposed a method of replacing and copying an unknown word token with a bilingual dictionary in neural MT.
They automatically constructed a translation dictionary from a training corpus using a word-alignment model (GIZA++), which finds a corresponding source word for each unknown target word token. 
They replaced the unknown word token with the corresponding word into which the source word was translated by the bilingual dictionary. 
\citet{yuan-briscoe2016} used a similar method for neural GEC.
Because our proposed method is performed as preprocessing, it can be used simultaneously with this replace-and-copy method.

Algorithms that rank words using co-occurrence are employed in many natural language processing tasks.
For example, \mbox{TextRank} \cite{mihalcea2004} uses PageRank \cite{brin_and_page1998} for keyword extraction.
TextRank constructs a word graph in which nodes represent words, and edges represent co-occurrences between words within a fixed window; TextRank then executes the PageRank algorithm to extract keywords.
Although this is an unsupervised method, it achieves nearly the same precision as one state-of-the-art supervised method \cite{hulth2003}.
\citet{kiso2011} used HITS \cite{kleinberg1999} to select seeds and create a stop list for bootstrapping in natural language processing.
They reported significant improvements over a baseline method using unigram frequency. 
Their graph-based algorithm was effective at extracting the relevance between words, which cannot be grasped with a simple unigram frequency.
In this study, we use HITS to retrieve co-occurring words from a training corpus to reduce noise in the source text.

\begin{algorithm}[t]
  \caption{HITS}
  \label{alg:something}
  \small
\begin{algorithmic}[1]
\Require hubness vector \bm{$i_0$}
\Require adjacency matrix \bm{$A$}
\Require iteration number $\tau$
\Ensure hubness vector \bm{$i$}  
\Ensure authority vector \bm{$p$}
\Function{HITS}{\bm{$i_0$}, \bm{$A$}, $\tau$}
  \State $\bm{i} \gets \bm{i_0}$
  \For{$t = 1, 2, ..., \tau$}
    \State $\bm{p} \gets \bm{A^{\mathrm{T}}} \bm{i}$
    \State $\bm{i} \gets \bm{A} \bm{p}$
    \State normalize \bm{$i$} and \bm{$p$} 
  \EndFor
  \State \Return \bm{$i$} and \bm{$p$}
\EndFunction
\end{algorithmic}
\label{tab:algohits}
\end{algorithm}

\section{Graph-based Filtering of OOV Words}
\subsection{Hubness and authority scores from HITS}
HITS, which is a web page ranking algorithm proposed by \citet{kleinberg1999}, computes hubness and authority scores for a web page (node) using the adjacency matrix that represents the web page's link (edge) transitions.
A web page with high authority is linked from a page with high hubness scores, and a web page with a high hubness score links to a page with a high authority score.
Algorithm \ref{tab:algohits} shows pseudocode for the HITS algorithm. Hubness and authority scores converge by setting the iteration number $\tau$ to a sufficiently large value.

\subsection{Vocabulary selection using HITS}
In this study, we create an adjacency matrix from a training corpus by considering a word as a node and the co-occurrence between words as an edge.
Unlike in web pages, co-occurrence between words is nonbinary; therefore, several co-occurrence measures can be used as edge weights.
Section 3.3 describes the co-occurrence measures and the context in which co-occurrence is defined.

The HITS algorithm is executed using the adjacency matrix created in the way described above.
As a result, it is possible to obtain a score indicating importance of each word while considering contextual information in the training corpus.

\begin{figure}[t]
  \small
  \begin{center}
    \includegraphics[scale = 0.3]{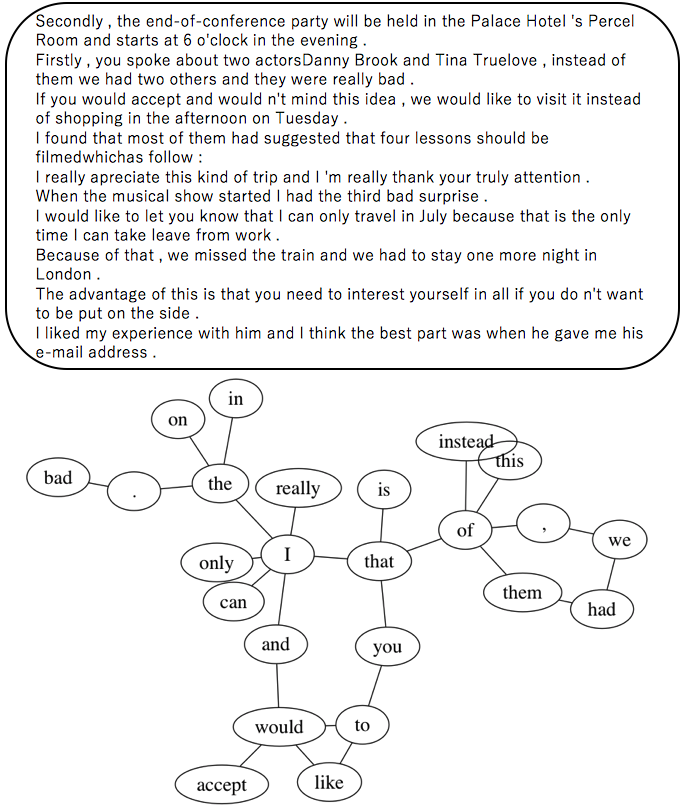}
    \caption{An example word graph created for ten sentences in the training corpus used for GEC\protect \footnotemark[2].}
    \label{tab:graphwords}
  \end{center}
  \vspace{-3mm}
\end{figure}

Figure \ref{tab:graphwords} shows a word graph example.
A word that obtains a high score in the HITS algorithm is considered to co-occur with a variety of words.
Figure \ref{tab:graphwords} demonstrates that second order co-occurrence scores (the scores of words co-occurring with words that co-occur with various words \cite{schutze98}) are also high.

In this study, words with high hubness scores are considered to co-occur with an important word, and low-scoring words are excluded from the vocabulary.
Using this method appears to generate a vocabulary that includes words that are more suitable for representing a context vector for encoder models.

\subsection{Word graph construction}
\begin{figure}[t]
  \small
  \begin{center}
    \includegraphics[scale = 0.23]{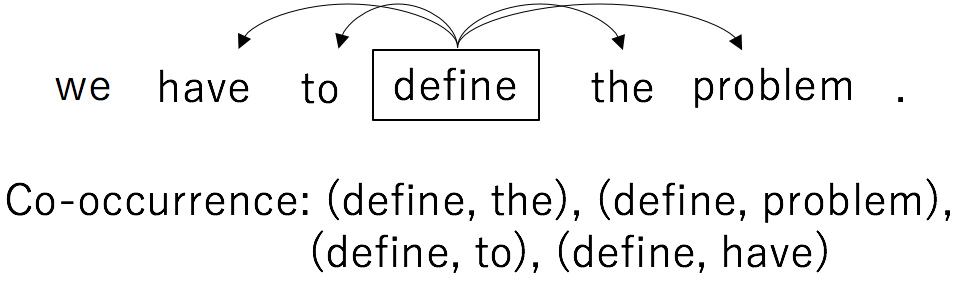}
    \caption{An example of co-occurrence context.}
    \label{tab:explain}
  \end{center}
  \vspace{-4mm}
\end{figure}

To acquire co-occurrence relations, we use a combination of each word and its peripheral words. 
Specifically, we combine the target word with surrounding words within window width $N$ and count the occurrences.
When defining the context in this way, because the adjacency matrix becomes symmetric, the same hubness and authority scores can be obtained. 
Figure \ref{tab:explain} shows an example of co-occurrence in which $N$ is set to two.
\footnotetext[2]{In this study, singleton words and their co-occurrences are excluded from the graph.}

We use raw co-occurrence frequency (Freq) and positive pointwise mutual information (PPMI) between words as the ($x, y$) element $A_{xy}$ of the adjacency matrix.
However, naive PPMI reacts sensitively to low-frequency words in a training corpus. 
To account for high-frequency, we weight the PMI by the logarithm of the number of co-occurrences and use PPMI based on this weighted PMI (Equation \ref{ppmi_equation}).
\begin{eqnarray}
  \label{freq_equation} \lefteqn{\hspace{-64mm}A_{xy}^{freq} = |x, y| }\\ 
  \label{ppmi_equation} A_{xy}^{ppmi}  = \mathrm{max}(0, \mathrm{pmi}(x, y) + \log_2|x, y|)  
\end{eqnarray}
Equation \ref{pmi_equation} is the PMI of target word $x$ and co-occurrence word $y$.
$M$ is the number of tokens of the combination, $|x, *|$ and $|*, y| $ are the number of token combinations when fixing target word $x$ and co-occurrence word $y$, respectively.
\begin{eqnarray}
  \label{pmi_equation} \mathrm{pmi}(x, y) = \log_2 \frac{M \cdot |x, y|}{|x, *||*, y|} 
\end{eqnarray}

\section{Machine Translation}
\subsection{Experimental setting}
In the first experiment, we conduct a Japanese-to-English translation using the Asian Scientific Paper Excerpt Corpus (ASPEC; \citeauthor{nakazawa}, \citeyear{nakazawa}). 
We follow the official split of the train, development, and test sets.
As training data, we use only the first 1.5 million sentences sorted by sentence alignment confidence to obtain a Japanese--English parallel corpus (sentences of more than 60 words are excluded).
Our training set consists of 1,456,278 sentences, development set consists of 1,790 sentences, and test set consists of 1,812 sentences.
The training set has 247,281 Japanese word types and 476,608 English word types.

The co-occurrence window width $N$ is set to two.
For combinations that co-occurred only once within the training corpus, we set the value of element $A_{xy}$ of the adjacency matrix to zero.
The iteration number $\tau$ of the HITS algorithm is set to 300.
As mentioned in Section 1, we only use the proposed method on the encoder side.

For this study's neural MT model\footnote[3]{\fontsize{7.5pt}{0cm}\selectfont \url{https://github.com/yukio326/nmt-chainer}}, we implement global dot attention \cite{luong2015b}. 
We train a baseline model that uses vocabulary that is determined by its frequency in the training corpus.
Vocabulary size is set to 100K on the encoder side and 50K on the decoder side.
Additionally, we conduct an experiment of varying vocabulary size of the encoder to 50K in the baseline and PPMI to investigate the effect of vocabulary size.
Unless otherwise noted, we conduct an analysis of the model using the vocabulary size of 100K.
The number of dimensions for each of the hidden and embedding layers is 512. 
The mini-batch size is 150. AdaGrad is used as an optimization method with an initial learning rate of 0.01.
Dropout is applied with a probability of 0.2.

For this experiment, a bilingual dictionary is prepared for postprocessing unknown words \cite{jean2015}. 
When the model outputs an unknown word token, the word with the highest attention score is used as a query to replace the unknown token with the corresponding word from the dictionary. 
If not in the dictionary, we replace the unknown word token with the source word (unk\_rep).
This dictionary is created based on word alignment obtained using fast\_align \cite{dyer2013} on the training corpus.

We evaluate translation results using BLEU scores \cite{papineni2002}. 

\begin{table}[t]
  \centering
  \small
  \begin{tabular}{l|rrr} \hline
    & baseline & HITS (Freq) & HITS (PPMI) \\ \hline \hline
   BLEU (50K) & 22.24 & - & 22.40 \\  \hline
   BLEU (100K) & 22.21 & 22.25 & \textbf{22.77} \\
   $p$-value & - & 0.35 & 0.01\\ \hline
  \end{tabular}
  \caption{BLEU scores for Japanese-to-English translation\protect \footnotemark[4]. The parentheses indicate vocabulary size of the encoder.}
  \label{tab:result1}
\end{table}
\footnotetext[4]{BLEU score for postprocessing (unk\_rep) improves by 0.46, 0.44, and 0.46 points in the baseline, Freq, and PPMI, respectively.}

\begin{table}[t]
  \centering
  \small
  \begin{tabular}{l||cc|cc} \hline
   & \multicolumn{2}{c|}{COMMON outputs} & \multicolumn{2}{c}{DIFF outputs}  \\
    & baseline & PPMI & baseline & PPMI\\ \hline
   BLEU & 22.33 & \textbf{22.98} & 21.44 & \textbf{21.98}\\
   \hline
  \end{tabular}
  \caption{BLEU scores of the COMMON and DIFF outputs.}
  \label{tab:incorexc}
\end{table}

\subsection{Results}
Table \ref{tab:result1} shows the translation accuracy (BLEU scores) and $p$-value of a significance test ($p < 0.05$) by bootstrap resampling \cite{koehn2004}.
The PPMI model improves translation accuracy by 0.56 points in Japanese-to-English translation, which is a significant improvement.

Next, we examine differences in vocabulary by comparing each model with the baseline.
Compared to the vocabulary of the baseline in 100K setting, Freq and PPMI replace 16,107 and 17,166 types, respectively; compared to the vocabulary of the baseline in 50K setting, PPMI replaces 4,791 types.

\begin{table*}[h]
  \centering
  \small
  \hspace*{-2em}
  \begin{tabular}{c|p{15cm}} \hline
    src &  有用 物質 の 分離 ・ 抽出 , 反応 性 向上 , 新 材料 創製 , 廃棄 物 処理 , 分析 等 の 分野 が ある 。\\ \hline\hline
    baseline & there are fields such as separation , extraction , extraction , improvement of new material creation , waste treatment , analysis , etc .\\\hline 
    Freq & there are separation and extraction of useful substances , the improvement of reactivity , new material creation , waste treatment and analysis .\\\hline
    PPMI & there are the fields such as separation and extraction of useful materials , the reaction improvement , new material creation , waste treatment , analysis , etc ... \\\hline
    ref &the application fields are separation and extraction of useful substances , reactivity improvement , creation of new products , waste treatment , and chemical analysis .\\\hline
  \end{tabular}
  \caption{An example of Japanese-to-English translation on a source sentence from COMMON.}
  \label{tab:resultexam}
\end{table*}

\subsection{Analysis}
According to Table \ref{tab:result1}, the performance of Freq is almost the same as that of the baseline.
When examining the differences in selected words in vocabulary between PPMI and Freq, we find that PPMI selects more low-frequency words in the training corpus compared to Freq, because PPMI deals with not only frequency but also co-occurrence.

The effect of unk\_rep is almost the same in the baseline as in the proposed method, which indicates that the proposed method can be combined with other schemes as a preprocessing step.

As a comparison of the vocabulary size 50K and 100K, the BLEU score of 100K is higher than that of 50K in PPMI.
Moreover, the BLEU scores are almost the same in the baseline.
We suppose that the larger the vocabulary size of encoder, the more noisy words the baseline includes, while the PPMI filters these words.
That is why the proposed method works well in the case where the vocabulary size is large.

To examine the effect of changing the vocabulary on the source side, the test set is divided into two subsets: COMMON and DIFF.
The former (1,484 sentences) consists of only the common vocabulary between the baseline and PPMI, whereas the latter (328 sentences) includes at least one word excluded from the common vocabulary.

Table \ref{tab:incorexc} shows the translation accuracy of the COMMON and DIFF outputs.
Translation performance of both corpora is improved.

In order to observe how PPMI improves COMMON outputs, we measure the similarity of the baseline and PPMI output sentences by counting the exact same sentences.
In the COMMON outputs, 72 sentence pairs (4.85\%) are the same, whereas 9 sentence pairs are the same in the DIFF outputs (2.74\%).
Surprisingly, even though it uses the same vocabulary, PPMI often outputs different but fluent sentences.

Table \ref{tab:resultexam} shows an example of Japanese-to-English translation.
The outputs of the proposed method (especially PPMI) are improved, despite the source sentence being expressed with common vocabulary; this is because the proposed method yielded a better encoder model than the baseline. 

\section{Grammatical Error Correction}
\subsection{Experimental setting}
The second experiment addresses GEC.
We combine the FCE public dataset \cite{yann2011}, NUCLE corpus \cite{dahlmeier2013}, and English learner corpus from the Lang-8 learner corpus \cite{mizumoto2011} and remove sentences longer than 100 words to create a training corpus. From the Lang-8 learner corpus, we use only the pairs of erroneous and corrected sentences. 
We use 1,452,584 sentences as a training set (502,908 types on the encoder side and 639,574 types on the decoder side).
We evaluate the models' performances on the standard sets from the CoNLL-14 shared task \cite{ng2014} using CoNLL-13 data as a development set (1,381 sentences) and CoNLL-14 data as a test set (1,312 sentences)\footnote[5]{We do not consider alternative answers suggested by the participating teams.}.
We employ $\mathrm{F_{0.5}}$ as an evaluation measure for the CoNLL-14 shared task.

\begin{table}[t]
  \centering
  \small
  \begin{tabular}{l||rr|rr} \hline
     & \multicolumn{2}{c|}{baseline} & \multicolumn{2}{c}{PPMI} \\
     & 50K  & 150K & 50K & 150K  \\ \hline
    Precision & 48.09 & 46.53 & \textbf{49.45} & 49.23  \\  \hline
    Recall & 8.30& 8.50 & 8.61 & \textbf{9.02} \\ \hline
    $\mathrm{F_{0.5}}$ & 24.55 & 24.55 & 25.37 & \textbf{26.03} \\ 
    \hline
  \end{tabular}
  \caption{$\mathrm{F_{0.5}}$ results on the CoNLL-14 test set\protect \footnotemark[5].}
  \label{tab:gec_result}
\end{table}

\begin{table}[t]
  \centering
  \small
  \begin{tabular}{l||rr|rr} \hline
   & \multicolumn{2}{c|}{COMMON outputs} & \multicolumn{2}{c}{DIFF outputs}  \\
    & baseline & PPMI & baseline & PPMI\\ \hline
   P & 48.26 & 60.07 & 9.40 & 17.32\\
   R & 0.01 & 0.01 & 0.01 & 0.02\\
   $\mathrm{F_{0.5}}$ & 0.04 & 0.04 & 0.04 & 0.08\\
   \hline
  \end{tabular}
  \caption{$\mathrm{F_{0.5}}$ of COMMON and DIFF outputs.}
  \label{tab:incorexc4gec}
\end{table}

We use the same model as in Section 4.1 as a neural model for GEC. The models' parameter settings are similar to the MT experiment, except for the vocabulary and batch sizes. In this experiment, we set the vocabulary size on the encoder and decoder sides to 150K and 50K, respectively. 
Additionally, we conduct the experiment of changing vocabulary size of the encoder to 50K to investigate the effect of the vocabulary size.
Unless otherwise noted, we conduct an analysis of the model using the vocabulary size of 150K.
The mini-batch size is 100.

\subsection{Result}
Table \ref{tab:gec_result} shows the performance of the baseline and proposed method.
The PPMI model improves precision and recall; it achieves a $\mathrm{F_{0.5}}$-measure 1.48 points higher than the baseline method. 

In setting the vocabulary size of encoder to 150K, PPMI replaces 37,185 types from the baseline; in the 50K setting, PPMI replaces 10,203 types.

\begin{table}[t]
  \centering
  \small
  \begin{tabular}{c|p{5.5cm}} \hline
    src &  Genetic refers the chance of inheriting a disorder or disease .\\ \hline\hline
    baseline & Genetic refers the chance of inheriting a disorder or disease .\\\hline 
    PPMI & Genetic refers \textbf{to} the chance of inheriting a disorder or disease .\\\hline
    gold &Genetic \textbf{risk} refers \textbf{to} the chance of inheriting a disorder or disease .\\\hline
  \end{tabular}
  \caption{An example of GEC using a source sentence from COMMON.}
  \label{tab:resultexam4gec}
\end{table}

\subsection{Analysis}
The $\mathrm{F_{0.5}}$ of the baseline is almost the same while the PPMI model improves the score in the case where the vocabulary size increases.
Similar to MT, we suppose that the PPMI filters noisy words.

As in Section 4.3, we perform a follow-up experiment using two data subsets: COMMON and DIFF, which contain 1,072 and 240 sentences, respectively.

Table \ref{tab:incorexc4gec} shows the accuracy of the error correction of the COMMON and DIFF outputs.
Precision increases by 11.81 points, whereas recall remains the same for the COMMON outputs.

In GEC, approximately 20\% of COMMON's output pairs differ, which is caused by the differences in the training environment.
Unlike MT, we can copy OOV in the target sentence from the source sentence without loss of fluency; therefore, our model has little effect on recall, whereas its precision improves because of noise reduction. 

Table \ref{tab:resultexam4gec} shows an example of GEC. 
The proposed method's output improves when the source sentence is expressed using common vocabulary.

\section{Discussion}
We described that the proposed method has a positive effect on learning the encoder.  
However, we have a question; what affects the performance?
We conduct an analysis of this question in this section.

First, we count the occurrence of the words included only in the baseline or PPMI in the training corpus.
We also show the number of the tokens per types (``Ave. tokens'') included only in either the baseline or PPMI vocabulary. 

\begin{table}[t]
  \centering
  \small
  \begin{tabular}{l||rr|rr} \hline
  & \multicolumn{2}{c|}{MT} & \multicolumn{2}{c}{GEC} \\ 
  & \multicolumn{1}{c}{baseline} & \multicolumn{1}{c|}{PPMI} & \multicolumn{1}{c}{baseline} & \multicolumn{1}{c}{PPMI} \\ \hline
  tokens &  52,700 & 27,364 & 126,884 & 70,003 \\
  Ave. tokens & 3.07 & 1.59 & 3.36 & 1.85 \\
  \hline
  \end{tabular}
  \caption{Number of words included only in either the baseline or PPMI vocabulary. }
  \label{tab:wordCount}
\end{table} 

The result is shown in Table \ref{tab:wordCount}.
We find that the proposed method uses low-frequency words instead of high-frequency words in the training corpus. 
This result suggests that the proposed method works well despite the fact that the encoder of the proposed method encounters more \verb|<unk>| than the baseline.
This is because the proposed method excludes words that may interfere with the learning of encoder-decoder models.

Second, we conduct an analysis of the POS of the words in GEC to find why increasing OOV improves the learning of encoder-decoder models. 
Specifically, we apply POS tagging to the training corpus and calculate the occurrence of the POS of the words only included in the baseline or PPMI.
We use NLTK as a POS tagger.

Table \ref{tab:posCount} shows the result.
It is observed that NOUN is the most affected POS by the proposed method and becomes often represented by \verb|<unk>|. 
NOUN words in the vocabulary of the baseline contain some non-English words, such as Japanese or Korean.
These words should be treated as OOV but the baseline fails to exclude them using only the frequency.
According to Table \ref{tab:posCount}, NUM is also affected by the proposed method.
NUM words of the baseline include a simple numeral such as ``\verb|119|'', in addition to incorrectly segmented numerals such as ``\verb|514&objID|''.
This word appears 25 times in the training corpus owing to the noisy nature of Lang-8.
We suppose that the proposed method excludes these noisy words and has a positive effect on training.

\begin{table}[t]
  \centering
  \begin{tabular}{l|rrr} \hline
   POS & \multicolumn{1}{c}{baseline} & \multicolumn{1}{c}{PPMI} & \multicolumn{1}{c}{ALL} \\ \hline
   NOUN & 92,693 & 44,472 & 4,644,478\\
   VERB & 11,066 & 10,099 & 3,597,895 \\
   PRON & 127 & 107 & 1,869,422 \\
   ADP & 626 & 685 & 1,836,193 \\
   DET & 128 & 202 & 1,473,391 \\
   ADJ & 13,855 & 12,270 & 1,429,056\\ 
   ADV & 2,032 & 1,688 & 931,763 \\
   PRT & 319 & 75 & 615,817 \\
   CONJ & 62 & 28 & 537,346 \\
   PUNCT & 110 & 11 & 223,573 \\
   NUM & 5,585 & 299 & 207,487 \\
   OTHER & 281 & 67 & 5,209 \\ \hline
   Total & 126,884 & 70,003 & 17,371,630 \\
   \hline
  \end{tabular}
  \caption{Number of the POS of words only included in the baseline or PPMI.}
  \label{tab:posCount}
\end{table}

\section{Conclusion}
In this paper, we proposed an OOV filtering method, which considers word co-occurrence information for encoder-decoder models. 
Unlike conventional OOV handling, this graph-based method selects the words that are more suitable for learning encoder models by considering contextual information. 
This method is effective for not only machine translation but also grammatical error correction.

This study employed a symmetric matrix (similar to skip-gram with negative sampling) to express relationships between words. 
In future research, we will develop this method by using vocabulary obtained by designing an asymmetric matrix to incorporate syntactic relations.

\section*{Acknowledgments}
We thank Yangyang Xi of Lang-8, Inc. for allowing us to use the Lang-8 learner corpus.
We also thank Masahiro Kaneko and anonymous reviewers for their insightful comments.

\bibliography{b4_ref}
\bibliographystyle{acl_natbib}

\end{document}